# T-GRAG: A Dynamic GraphRAG Framework for Resolving Temporal Conflicts and Redundancy in Knowledge Retrieval

Dong Li*
arvinlee826@gmail.com
School of Mathematics, Harbin
Institute of Technology
Harbin, China

Yichen Niu*
School of Astronautics, Harbin
Institute of Technology
Harbin, China
niu62@stu.hit.edu.cn

Ying Ai
School of Astronautics, Harbin
Institute of Technology
Harbin, China

Xiang Zou
School of Mathematics, Harbin
Institute of Technology
Harbin, China

Biqing Qi†
Shanghai Artificial Intelligence
Laboratory
Shanghai, China

Jianxing Liu†
School of Astronautics, Harbin
Institute of Technology
Harbin, China

## Abstract

Large language models (LLMs) have demonstrated strong performance in natural language generation but remain limited in knowledge-intensive tasks due to outdated or incomplete internal knowledge. Retrieval-Augmented Generation (RAG) addresses this by incorporating external retrieval, with GraphRAG further enhancing performance through structured knowledge graphs and multi-hop reasoning. However, existing GraphRAG methods largely ignore the temporal dynamics of knowledge, leading to issues such as temporal ambiguity, time-insensitive retrieval, and semantic redundancy. To overcome these limitations, we propose Temporal GraphRAG (T-GRAG), a dynamic, temporally-aware RAG framework that models the evolution of knowledge over time. T-GRAG consists of five key components: (1) a Temporal Knowledge Graph Generator that creates time-stamped, evolving graph structures; (2) a Temporal Query Decomposition mechanism that breaks complex temporal queries into manageable sub-queries; (3) a Three-layer Interactive Retriever that progressively filters and refines retrieval across temporal subgraphs; (4) a Source Text Extractor to mitigate noise; and (5) a LLM-based Generator that synthesizes contextually and temporally accurate responses. We also introduce Time-LongQA, a novel benchmark dataset based on real-world corporate annual reports, designed to test temporal reasoning across evolving knowledge. Extensive experiments show that T-GRAG significantly outperforms prior RAG and GraphRAG baselines in both retrieval accuracy and response relevance under temporal constraints, highlighting the necessity of modeling knowledge evolution for robust long-text question answering. Our code is publicly available on the T-GRAG.

## 1 Introduction

With the significant breakthroughs achieved by large language models [17], such as GPT [2], LLAMA [4] in the field of natural language processing (NLP), text generation tasks based on these models have demonstrated remarkable capabilities in a variety of application scenarios. However, in complex tasks that require specific domain knowledge or external information, these models are often limited by outdated knowledge and insufficient content accuracy [21]. To address this issue, the Retrieval-Augmented Generation (RAG) method has emerged [6]. In recent years, the RAG framework has continued to evolve, giving rise to various variants. Among them, GraphRAG [5], which constructs a graph index and performs retrieval at the node level, has broken through the limitations of traditional RAG methods based on text similarity. It has shown outstanding performance in multi-hop reasoning [9, 12] and long-text question-answering tasks [11], and has thus attracted widespread attention. However, in real-world applications, knowledge is not static but continuously evolves over time [8, 13, 14]. The incremental update mechanism of the existing GraphRAG is merely a simple stacking of graph information, without considering the temporal evolution characteristics of the graph [16]. As the knowledge base continues to expand, GraphRAG faces three key challenges.

**Challenge 1: Temporal ambiguity in knowledge modeling leads to indexing errors.** In real-world scenarios, knowledge evolves over time. An entity may exhibit different attributes at different time points—for instance, a company's net cash flow, product delivery volume, or organizational structure may vary across years [15]. However, existing GraphRAG methods often disregard such temporal characteristics when constructing the knowledge graph or building indexable embeddings. As a result, time-specific facts are merged under a single entity node without explicit temporal separation, leading to temporal ambiguity in the representation. This modeling oversight prevents GraphRAG from distinguishing between facts that are semantically similar but temporally distinct. For example, as illustrated in Figure 1, multiple automobile delivery records associated with Audi Group are included in the graph, but their respective years are not encoded. Consequently, when the model encounters a knowledge entry such as "1.8 million units delivered," it cannot determine whether this figure refers to 2022 or 2023, making it difficult to provide accurate and temporally grounded

---

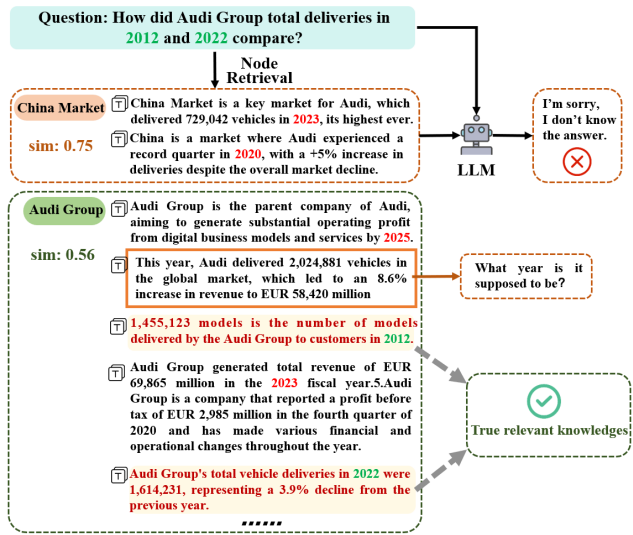

Figure 1: An example of a GraphRAG retrieval failure due to temporal evolution of knowledge and node redundant knowledge.

responses. This issue originates from the construction phase, where time-awareness is lacking in both graph structure and node-level embedding.

**Challenge 2: Time-insensitive retrieval introduces irrelevant results.** Even when temporal information is available in the knowledge graph, current GraphRAG retrieval mechanisms typically fail to leverage it effectively during the search process. Specifically, the retrieval stage often prioritizes semantic similarity while ignoring the temporal constraints embedded in the user query. As a result, the system may retrieve information that is topically relevant but temporally mismatched. For instance, as shown in Figure 1, when a user asks for Audi Group's vehicle deliveries in 2012 or 2022, the system may still retrieve data from 2023 if it shares similar semantic context. This occurs because the model does not treat the query's temporal reference as a hard constraint or retrieval signal. Such behavior leads to inaccurate or misleading outputs, especially in domains where temporal consistency is critical (e.g., financial reporting, historical analysis). Unlike Challenge 1, this issue arises at the retrieval stage, where the system fails to align the search results with the query's temporal intent.

**Challenge 3: Redundant information at the same time point disrupts semantic relevance in retrieval.** Even when the knowledge is correctly aligned with a specific time point, challenges remain due to semantic redundancy within individual nodes. A single entity at a given timestamp may be associated with a wide range of heterogeneous information—some of which may be irrelevant to the user's query. When such diverse knowledge is densely embedded into a single node, it can dilute the node's semantic representation and compromise retrieval accuracy. For example, as shown in Figure 1, the node representing Audi Group in 2022 may include multiple types of facts: vehicle delivery statistics, financial performance, organizational changes, and more. While all of this information belongs to the same time frame, not all of it is relevant to a specific query. Suppose a user is interested only in Audi's vehicle delivery volume in 2022. If the node's representation is dominated or distorted by unrelated financial information, the resulting embedding may appear less similar to the query. As a consequence, the node may be ranked lower or even excluded during retrieval.

In order to address the above challenge, we propose **Temporal GraphRAG (T-GRAG)**, a general, dynamic, and precise temporal GraphRAG method. The T-GRAG mainly consists of five plug-and-play modules: Temporal Knowledge Graph Generator, Temporal Query Decomposition, Three-layer Interactive Retriever, Valid Source Text Extractor, and Augmented Generator based on LLM.

In the indexing phase, the temporal knowledge graph generator constructs a time-aware knowledge graph in which each node, edge, and associated fact is annotated with temporal attributes. This design enables fine-grained differentiation of knowledge across different time periods. During the retrieval phase, the time query decomposer first breaks down complex queries containing multiple temporal constraints into simpler sub-queries, each focusing on a single time condition, thereby reducing retrieval complexity. These sub-queries are then processed by a Three-layer Interactive Retriever, which consists of a temporal subgraph retriever, a coarse-grained node retriever, and a fine-grained knowledge retriever. Specifically, the temporal subgraph retriever extracts subgraphs at the relevant time points, preventing retrieval errors and conflicts arising from temporal evolution. The coarse-grained node retriever encodes both the query and the nodes within the subgraph, calculates their semantic similarity, and selects the top-n most relevant nodes as candidates. Building on this, the fine-grained knowledge retriever matches the query to knowledge associated with the candidate nodes and identifies the top-k most relevant pieces of information, while marking the corresponding nodes as relevant. Based on the selected nodes, the valid source text extractor retrieves the top-t supporting texts, effectively filtering out noise introduced by redundant or outdated knowledge. In the enhancement generation phase, the retrieved subgraph information and source texts are fed into a large language model (LLM)-augmented generator, which integrates these inputs to produce the final answer.

To evaluate the performance of RAG methods in handling long-text question answering tasks with temporal knowledge bases, we constructed a dataset called **Time-LongQA Dataset**. The dataset is derived from the company's annual reports, as these reports contain a large amount of temporal evolution knowledge, and most of the content is unseen by LLMs. The questions cover four types of time constraints, single time, dual time, multiple time ($\geq$ 3 time points), and no time. We designed an automated process for generating and manually validating QA pairs, extracting 2,292 question-answer pairs. Based on this dataset, we conducted multiple experiments on GraphRAG, Vanilla RAG, and T-GRAG, demonstrating the effectiveness of T-GRAG in temporal long-text question answering tasks.

In summary, our contributions are as follows:

• **Temporal GraphRAG (T-GRAG)**: We propose a novel dynamic retrieval-augmented generation framework that constructs and utilizes temporal knowledge graphs. This design addresses the challenge of temporal ambiguity in traditional GraphRAG systems, enabling time-sensitive indexing and retrieval.

• **Temporal Query Decomposition (TQD)**: We introduce a query decomposition strategy that splits complex queries with multiple time constraints into simpler single-time sub-queries. This effectively avoids inference confusion across time periods and narrows the retrieval space, reducing query complexity.

• **Three-layer Interactive Retriever**: We design a hierarchical retriever consisting of a temporal subgraph retriever, coarse-grained node retriever, and fine-grained knowledge retriever. This structure mitigates retrieval errors caused by temporal mismatch and node-level semantic redundancy, improving precision and granularity of retrieval.

• **Time-LongQA Dataset**: We construct a high-quality long-text QA dataset with diverse temporal constraints derived from real-world company reports. The dataset reflects knowledge evolution across time and enables fair and challenging evaluation of temporal retrieval systems. T-GRAG achieves state-of-the-art performance on this benchmark.

## 2 Temporal GraphRAG

In this section, we provide a detailed introduction to the five key modules of T-GRAG: Temporal Knowledge Graph Generator, Temporal Query Decomposition, Three-layer Interactive Retriever, Valid Source Text Extractor, and LLM Augmented Generator. The overall pipeline of T-GRAG is illustrated in Fig.2.

### 2.1 Temporal Knowledge Graph Generator

To meet the temporal evolution requirements of textual knowledge bases, we have designed a generator with dynamic updating capabilities. We first divide the temporal source text knowledge base $D^T$ into multiple time periods $\{D^{t_1}, D^{t_2}, ..., D^{t_n}\}$, facilitating the modeling of temporal evolution. Then, we split the text in each time period $D^{t_i}$ into fixed-sized text blocks $\{d_1^{t_i}, d_2^{t_i}, ..., d_c^{t_i}\}$ which serve as the processing units. Next we use LLM to extract graph information. First, we identify all entities in the text blocks, including the entity's name, type, and description. Then, we identify the relationships between related entities, clarifying the relationship type and description between source and target entities. Since the semantic information of the knowledge graph mainly resides in the descriptions of entities and relationships, we define these descriptions as knowledge. Meanwhile, we introduce temporal attributes into entities and relationships and associate the corresponding timestamp $t_i$ with knowledge $k$ to distinguish knowledge of the same entity at different time points. During the process of updating the graph, knowledge with the latest timestamp is continuously added to entities and relationships, avoiding confusion with past knowledge. Finally, we construct temporal dynamic graph knowledge base $TG = (\mathcal{E}, \mathcal{R})$. Each entity $e_i^{T_i} \in \mathcal{E}$ is defined as

$$e_i^{T_i} = \{k_1^{t_i}, k_2^{t_i}, ..., k_{m_{t_i}}^{t_i} \mid t_i \in T_i\} \quad (1)$$

where $T^i$ denote the time attribute of entity $e_i$, $k_{m_{t_i}}^{t_i}$ denote knowledge of $e_i$ occurring at time $t_i$.

---

**Algorithm 1** Temporal GraphRAG (T-GRAG)

1: **Input:** Temporal text knowledge base $D^T$, query $Q$
2: **Output:** Final answer Answer$_{\text{final}}$
3: **Step 1: Temporal Knowledge Graph Generation**
4: **for** each time period $D^{t_i}$ in $D^T$ **do**
5:     Split $D^{t_i}$ into text blocks $\{d_1^{t_i}, ..., d_c^{t_i}\}$
6:     **for** each block $d_j^{t_i}$ **do**
7:         Extract entities and relations with LLM
8:         Add temporal attributes and knowledge to graph $TG$
9:     **end for**
10: **end for**
11: **Step 2: Temporal Query Decomposition**
12: **if** Query $Q$ has temporal constraints **then**
13:     Decompose $Q$ into subqueries $\{q^{t_1}, ..., q^{t_q}\}$ by LLM
14: **else**
15:     Subqueries = $\{Q\}$
16: **end if**
17: **Step 3: Three-layer Interactive Retriever**
18: **for** each subquery $q^{t_i}$ **do**
19:     **3.1 Temporal Subgraph Retrieval**
20:     Retrieve subgraph $G_{t_i}$ for $t_i$ from $TG$
21:     Generate node embeddings $z_e$ for $G_{t_i}$
22:     **3.2 Node Retrieval**
23:     Generate query embedding $z_q$ for $q^{t_i}$
24:     Retrieve top-n relevant nodes $E_{\text{candidate}}$ by cosine similarity
25:     **3.3 Knowledge Retrieval**
26:     Retrieve top-k relevant knowledge $K_{\text{valid}}$ from $E_{\text{candidate}}$
27:     Identify valid nodes $E_{\text{valid}}$
28:     **Step 4: Source Text Extraction**
29:     Score and select top-t source texts $D_{\text{valid}}$
30:     **Step 5: LLM Augmented Generation**
31:     Generate answer $an^{t_i} = \text{LLM}(q^{t_i}, K_{\text{valid}}, D_{\text{valid}})$
32:     Store $(q^{t_i}, an^{t_i})$ in All_answers
33: **end for**
34: **Final Answer Generation**
35: Answer$_{\text{final}}$ = LLM($Q, \sum[q^{t_i}, an^{t_i}]$)
36: **return** Answer$_{\text{final}}$

---

### 2.2 Temporal Query Decomposition

Traditional GraphRAG retrieval typically provides LLMs with all relevant temporal information at once, but mixing knowledge from multiple time points may lead to reasoning confusion [20]. Additionally, if a query involves multiple temporal constraints, directly retrieving a large-scale temporal subgraph can cause a sharp increase in computational complexity and introduce unnecessary noise.

To address these issues, we introduce the Temporal Query Decomposer (TQD), leveraging the strong temporal semantic awareness of LLMs [10] to analyze, reason, and decompose temporal constraints in queries. Specifically, given an original query $Q$, TQD first utilizes the LLM to identify and parse its temporal constraints. If the query contains no temporal constraints, the original query remains unchanged, implying that the question requires knowledge spanning all time periods. However, if the query includes multiple temporal constraints, the LLM's reasoning capability is used to decompose Q into multiple independent subqueries $\{q^{t_1}, q^{t_2}, ..., q^{t_q}\}$, where each subquery corresponds to a single time point.

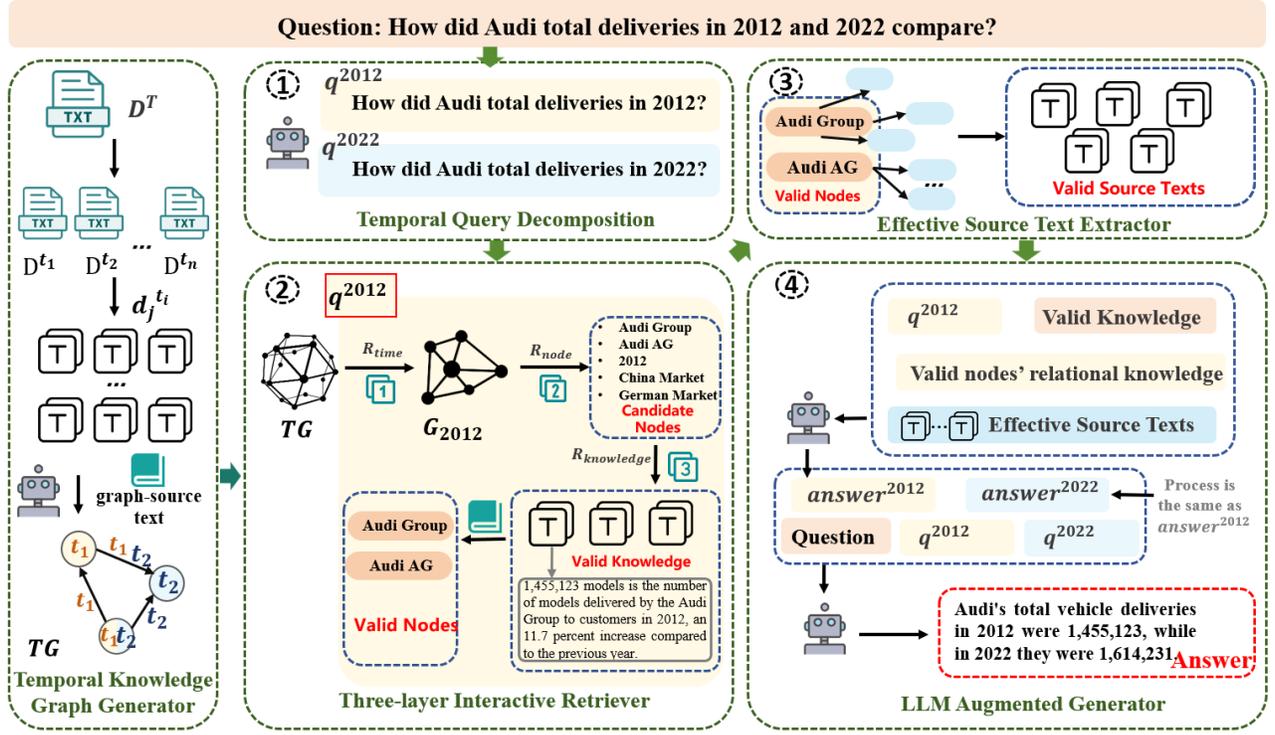

Figure 2: The the overall framework of T-GRAG.

### 2.3 Three-layer Interactive Retriever

To effectively address retrieval errors caused by the evolution of temporal knowledge and redundant knowledge in nodes, we propose a Three-layer Interactive Retriever $\mathcal{R}$. This retriever consists of temporal subgraph retriever, coarse-grained node retriever, and fine-grained knowledge retriever.

**Temporal subgraph retriever ($R_{time}$):** To avoid interference from irrelevant time periods in the graph knowledge and prevent knowledge conflicts, $R_{time}$ extracts subgraphs corresponding to the time attributes in the query from the knowledge graph, ensuring that each entity and relationship within the subgraph contains only knowledge relevant to that specific time point. Specifically, for a sub-temporal query $q^{t_i}$ with a timestamp $t_i$, we retrieve the temporal subgraph $G_{t_i}$ from the temporal knowledge graph $TG$, where the nodes and relationships in $G_{t_i}$ only contain knowledge from the time point $t_i$. Each entity that only contains knowledge of time $t_i$ is defined as $e_i^{t_i} \in G_{t_i}$.

$$e_i^{t_i} = \{k_1^{t_i}, k_2^{t_i}, ..., k_{m_{t_i}}^{t_i}\} \quad (2)$$

Additionally, we use a pre-trained embedding model (EM) to generate node embeddings for the $G_{t_i}$. Specifically, We encode all the knowledge in the nodes to generate the node embedding library $z_n$ for $G_{t_i}$.

$$z_e = \text{EM}(e_i^{t_i}) \in \mathbb{R}^d \quad (3)$$

where $d$ denotes the dimension of the output vector.

**Coarse-grained node retriever ($\mathcal{R}_{node}$) + Fine-grained knowledge retriever ($\mathcal{R}_{knowledge}$):** The traditional GraphRAG retrieval method performs retrieval only at the node level, which is ineffective at distinguishing the internal knowledge differences within the nodes. Since nodes may contain a large amount of redundant knowledge, simple node-level retrieval struggles to effectively differentiate between valuable information and redundant content, resulting in a decline in retrieval performance. Therefore, we propose an interactive method that combines the $\mathcal{R}_{node}$ retriever with the $\mathcal{R}_{knowledge}$ retriever. The specific process is as follows:

First, we apply the same embedding strategy to subquery $q^{t_i}$ to ensure consistent processing of textual information.

$$z_q = \text{EM}(q^{t_i}) \in \mathbb{R}^d \quad (4)$$

$\mathcal{R}_{node}$: Subsequently, to identify the candidate nodes relevant to the query, we use the nearest neighbor search method [1], where cosine similarity is employed to measure the similarity between the query embedding and node embeddings. The top-n most relevant nodes are selected as the candidate nodes $E_{candidate}$ for retrieval.

$$E_{candidate} = \text{argtopn}_{e \in E} \cos(z_q, z_e) \quad (5)$$

The argtopn operation retrieves the top-n elements based on this similarity, providing a set of $E_n$ considered most relevant to the query.

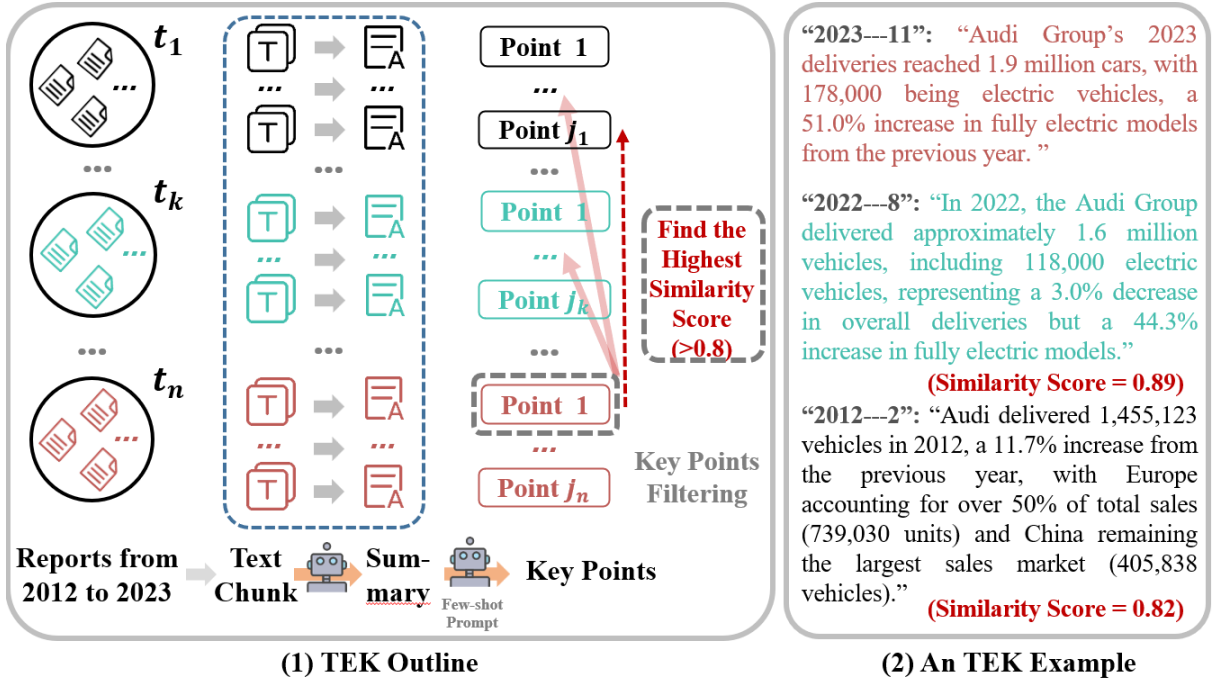

Figure 3: The left side shows the pipeline of finding TEK, while the right side presents an example of TEK.

$\mathcal{R}_{knowledge}$: Next, we will extract valid knowledge from the candidate nodes. Specifically, for each piece of knowledge within the candidate nodes, we generate embeddings individually, employing the same embedding strategy.

$$z_k = \mathbf{EM}(k^{t_i}) \in \mathbb{R}^d \mid k^{t_i} \in E^n \quad (6)$$

Then, cosine similarity is used to measure the distance between query and knowledge embeddings, selecting the top-k most similar knowledge blocks as the valid knowledge $K_{valid}$.

$$K_{valid} = \text{argtopk}_{e \in E} \cos(z_q, z_k) \quad (7)$$

Ultimately, we will consider nodes containing $K_{valid}$ as valid nodes $E_{valid}$

### 2.4 Valid Source Text Extractor

Previous GraphRAG retrieval methods offer significant advantages by leveraging the topological structure of graphs and refining the retrieval granularity to the node level. However, the knowledge encapsulated within graph nodes and relationships may still be incomplete relative to the original text. To address this, we identify the most effective source texts based on the effective nodes extracted by the three-layer retriever and their corresponding topological information.

Specifically, we score the candidate source texts according to the connectivity of the effective nodes and their neighboring nodes. The more 1-hop neighboring nodes a text block contains, the higher its score. Finally, the top-t source texts with the highest scores are selected as the final valid source texts $D_{valid}$.

### 2.5 LLM Augmented Generator

The valid graph information consists of the top-k relevant knowledge $K_{valid}$ extracted from three layers of interactive knowledge retrievers, as well as the relational knowledge $R_{valid}$ associated with valid nodes $E_{valid}$. This enables the LLM to understand the topological information of retrieved subgraphs. Simultaneously, the valid text information $D_{valid}$ includes the top-t relevant source texts extracted by the valid source text retriever, ensuring that the LLM has access to complete semantic context. The generator enhances the knowledge between these two perspectives to produce accurate answers to the query.

$$an^{t_i} = LLM(q^{t_i} + K_{valid} + R_{valid} + D_{valid}) \quad (8)$$

Ultimately, the original query, along with all sub-temporal queries $q^{t_i}$ and their corresponding answers $an^{t_i}$, aare fed into the LLM to generate a comprehensive final answer.

$$Answer_{final} = LLM(Q + \sum_{t_i \in \{t_1,...,t_q\}} [q^{t_i}, an^{t_i}]) \quad (9)$$

## 3 Time-LongQA Dataset

To construct a dataset of temporal evolution long texts while ensuring that most of the data is unseen by the LLM, we selected the publicly available Audi annual reports from 2012 to 2023 as the base corpus. Using the MinerU tool [18], we converted the source texts from PDF format to Markdown format, retaining only the textual content. Subsequently, we designed an automated temporal question-answer (QA) pairs construction process, which includes 1,538 single-time-constrained questions($Q_{Single}$), 524 dual-time-constrained questions($Q_{Dual}$), 113 multi-time-constrained(≥ 3 time points) questions($Q_{Multi}$), and 117 non-time-constrained questions($Q_{Non}$).The prompts for constructing dataset is detailed in the Appendix C.

### 3.1 QA-pair Generation and Validation

First, we divided each annual report into text blocks of 2,000 tokens. For $Q_{Dual}$ and $Q_{Multi}$, we developed a fully automated pipeline. This pipeline first identifies **temporal evolution knowledge (TEK)** within the temporal long texts and then generates questions involving two or more time points based on the extracted TEK. For $Q_{Single}$ and $Q_{Non}$, we utilized GPT-4 [2] to generate the corresponding QA pairs. Finally, all generated QA pairs underwent manual validation, with particular attention paid to avoiding potential temporal conflicts, especially when dealing with $Q_{Non}$ (such as changes in Audi's CEO).

### 3.2 How to Find TEK

Inspired by TCELongBench [24], we propose a TEK extraction process based on LLMs. The pipeline and example are shown in the figure3. The process is as follows: First, we generate summary representations for all text blocks at each timestamp $t_i$ to extract their core content. Next, we use few-shot prompt to derive key points from each text block's summary. Subsequently, starting from the key points of the most recent timestamp, we use cosine similarity search to find the most similar key points in each previous timestamp with a similarity score above a threshold. Finally, we found multiple TEK in a large number of annual reports.

## 4 Experiments

### 4.1 Experimental Setup

**Baselines**. We compare our proposed T-GRAG with three types of baseline methods: Standard LLMs (Base LLM), Text Retrieval-Augmented LLMs (Vanilla RAG) [7], and Graph Retrieval-Augmented LLMs (GraphRAG) [5].

- **Standard LLMs**: This setting evaluates whether a language model can answer questions solely based on its internal knowledge, without access to any external data.
- **Vanilla RAG**: We use the original textual corpus as the knowledge source, and employ a dense retriever to retrieve relevant textual passages. The retrieved content is then provided as additional context to augment the LLM for question answering.
- **GraphRAG**: We adopt the nano-GraphRAG pipeline [5] as the GraphRAG baseline. In this setting, a graph-structured knowledge base is first constructed from the original text. Retrieval is then performed over this graph to obtain semantically relevant subgraphs or nodes. The retrieved graph-based content, along with the original text, is provided as context to assist the LLM in generating accurate answers. Nano-GraphRAG also supports incremental updates to the graph, enabling dynamic knowledge integration.

For all baseline categories, we evaluate three LLM backbones: LLaMA-3.1-70B-Chat [4], Qwen2.5-72B-Instruct [22], and LLaMA-3.1-Nemotron-70B-Instruct-HF [19].

**Evaluation Metrics**. We adopt language model-based metrics to evaluate the quality of model responses, using Qwen2.5-72B to assess their consistency with ground-truth answers. Each sample is independently evaluated three times by the model, and a majority voting strategy [3] is applied: the answer is considered correct if at least two out of three evaluations deem it correct. The final LLMscore is measured by answer accuracy. The prompts used for LLM-based evaluation follow the DocBench protocol [25], with more details provided in Appendix A.1.

**Parameter Configuration**. For all RAG-based methods, we employ **stella-en-1.5B-v5** [23] as the embedding model for dense retrieval. The retrieval strategy over source text is parameterized by the product of chunk size and top-$t$, with chunk size set to 1000 and top-$t$ set to 5 in our main experiments. For GraphRAG and T-GRAG, the total number of retrieved context tokens is limited to 1600 tokens. For T-GRAG specifically, we set the number of candidate nodes ($n$) to 30, and select the top-$k$ most relevant knowledge units, with $k = 15$, based on the graph retrieval scores.

### 4.2 Overall Performance

In this section, we perform a multi-dimensional comparison and analysis of the overall performance results in Table 1.

**T-GRAG vs Others.** Among all LLM backbones, T-GRAG significantly outperforms all baselines. In particular, the improvement of T-GRAG in complex temporal constraints is more significant. On $Q_{Single}$, T-GRAG's score improved by an average of 22.28 compared to GraphRAG, and it improved by 34.05 and 47.2 on $Q_{Dual}$ and $Q_{Multi}$. This is because complex temporal constraints often introduce more temporal redundant knowledge, reducing the accuracy of the node retrieval. Experimental results robustly demonstrate T-GRAG's effectiveness in mitigating temporal retrieval errors. Additionally, for $Q_{Non}$, T-GRAG also outperforms both Vanilla RAG and GraphRAG, further validating the capability of its three-layer interactive retrieval framework to enhance retrieval accuracy in non-temporal tasks. The performance of Base LLMs is notably poor, which demonstrates that the necessary knowledge to answer these questions is not inherently contained within the LLM. The accuracy of the responses primarily depends on the ability to effectively retrieve relevant text.

**Graph RAG vs Vanilla RAG.** Compared to Vanilla RAG, which directly retrieves from the source text, GraphRAG features a more complex pipeline and incurs higher computational costs. However, our findings indicate that in certain cases, the performance of Graph RAG is worse than Vanilla RAG. This observation substantiates our argument in the introduction that the node embedding-based retrieval mechanism in traditional GraphRAG is less effective for complex long-text scenarios, particularly in queries requiring localized information. The primary reason lies in the excessive redundancy of node knowledge, which impairs the retriever's ability to accurately extract the relevant local information. Our ablation study "Impact of interactive retrieval between $\mathcal{R}_{node}$ and $\mathcal{R}_{knowledge}$" further demonstrates that the retrieval framework in T-GRAG effectively mitigates this issue.

Table 1: Model performance on Time-LongQA dataset comparing base LLMs, text retrieval augmented LLMs (Vanilla RAG), graph retrieval augmented LLMs (Graph RAG), and T-GRAG. In the case of the same problem type and the same LLM backbone, the increment of T-GRAG compared to the SOTA method will be displayed in red font.

|  | Model | Single-Time | Dual-Time | Multi-Time | Non-Time |
|---|---|---|---|---|---|
| **Base LLMs** | LLama-3.1-70b | 9.36 | 9.13 | 12.05 | 9.00 |
|  | Qwen-2.5-72b | 2.53 | 4.97 | 3.37 | 3.50 |
|  | Nemotron-3.1-70b | 12.79 | 10.82 | 11.93 | 5.50 |
|  | Llama3.2-8b-instruct | 7.58 | 7.88 | 8.63 | 6.34 |
| **Vanilla RAG** | LLama-3.1-70b | 59.36 | 35.11 | 30.97 | 64.10 |
|  | Qwen-2.5-72b | 64.69 | 50.95 | 46.01 | 68.37 |
|  | Nemotron-3.1-70b | 59.94 | 47.24 | 36.28 | 69.23 |
|  | Llama3.2-8b-instruct | 45.18 | 21.94 | 16.81 | 52.99 |
| **Graph RAG** | LLama-3.1-70b | 50.25 | 20.03 | 14.15 | 67.52 |
|  | Qwen-2.5-72b | 62.28 | 37.91 | 33.62 | 80.34 |
|  | Nemotron-3.1-70b | 59.10 | 30.72 | 24.77 | 73.50 |
|  | Llama3.2-8b-instruct | 35.95 | 9.16 | 8.84 | 53.84 |
| **T-GRAG** | LLama-3.1-70b | 78.15$_{18.79↑}$ | 58.77$_{23.66↑}$ | 69.02$_{38.05↑}$ | 71.79$_{4.27↑}$ |
|  | Qwen-2.5-72b | 78.15$_{13.46↑}$ | 63.74$_{12.76↑}$ | 69.91$_{23.90↑}$ | 81.19$_{0.85↑}$ |
|  | Nemotron-3.1-70b | 82.18$_{22.24↑}$ | 68.32$_{21.07↑}$ | 75.22$_{38.94↑}$ | 76.92$_{3.87↑}$ |
|  | LLama3.2-8b-instruct | 62.93$_{17.75↑}$ | 48.85$_{26.91↑}$ | 53.98$_{37.17↑}$ | 58.97$_{5.98↑}$ |

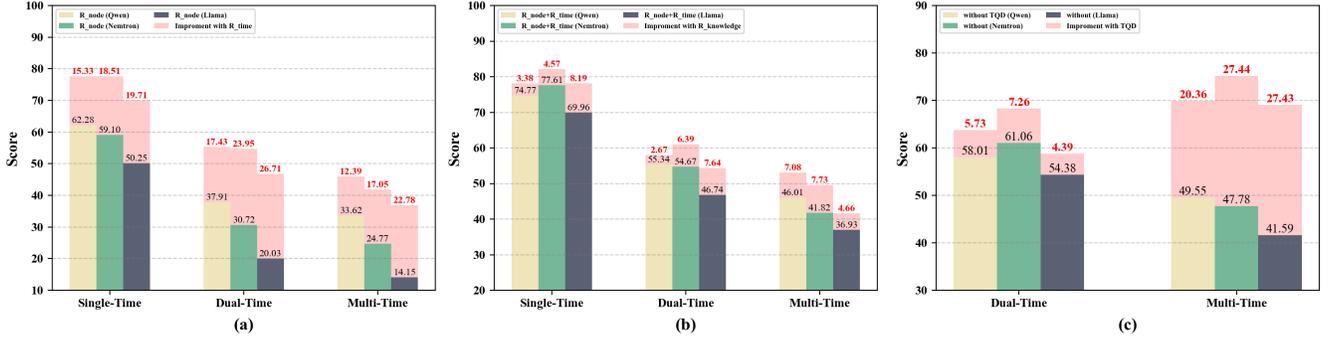

Figure 4: These three ablation experiments are conducted on three problem types: $Q_{Single}$, $Q_{Dual}$ and $Q_{Multi}$, using the same three LLM models as in the main experiment. The incremental parts are represented by pink bar charts and red font. (a) Results comparsion between $\mathcal{R}_{node}$ and $\mathcal{R}_{node}$ +$\mathcal{R}_{time}$. (b) Results comparsion between $\mathcal{R}_{node}$ +$\mathcal{R}_{time}$ and $\mathcal{R}_{node}$ +$\mathcal{R}_{time}$ + $\mathcal{R}_{knowledge}$. (c) Results comparsion between without TQD and with TQD.

## 4.3 Ablation Study

*4.3.1 How important is Three-layer Interactive Retriever?* To effectively verify the effectiveness of our retrieval mechanism, we conduct two sets of ablation experiments to investigate the effectiveness of T-GRAG's core components: (1) the impact of the temporal subgraph retriever $\mathcal{R}_{time}$ on answering time-sensitive queries; and (2) the contribution of the interactive retrieval mechanism between $\mathcal{R}_{node}$ and $\mathcal{R}_{knowledge}$ in enhancing retrieval accuracy and robustness. All experiments are conducted using the same backbone LLM (Qwen2.5-72B-Instruct) and retrieval configuration as in the main setup. Detailed performance comparisons are illustrated in Figure 4.

**Effect of Temporal Retriever $\mathcal{R}_{time}$.** To assess the role of temporal awareness, we compare two variants: (a) a standard GraphRAG pipeline using only $\mathcal{R}_{node}$, and (b) a temporal-aware version incorporating both $\mathcal{R}_{node}$ and $\mathcal{R}_{time}$. As shown in Figure 4(a), integrating $\mathcal{R}_{time}$ improves the overall answer accuracy by 19.31% on temporal queries. This demonstrates that incorporating temporal subgraph retrieval effectively reduces noise introduced by outdated or temporally irrelevant content. By filtering information from unrelated time slices, $\mathcal{R}_{time}$ mitigates conflicts caused by temporal evolution, thereby enhancing retrieval precision.

**Effect of Interactive Retrieval Between $\mathcal{R}_{node}$ and $\mathcal{R}_{knowledge}$.** We further examine whether the addition of $\mathcal{R}_{knowledge}$ — which retrieves fine-grained knowledge units associated with candidate

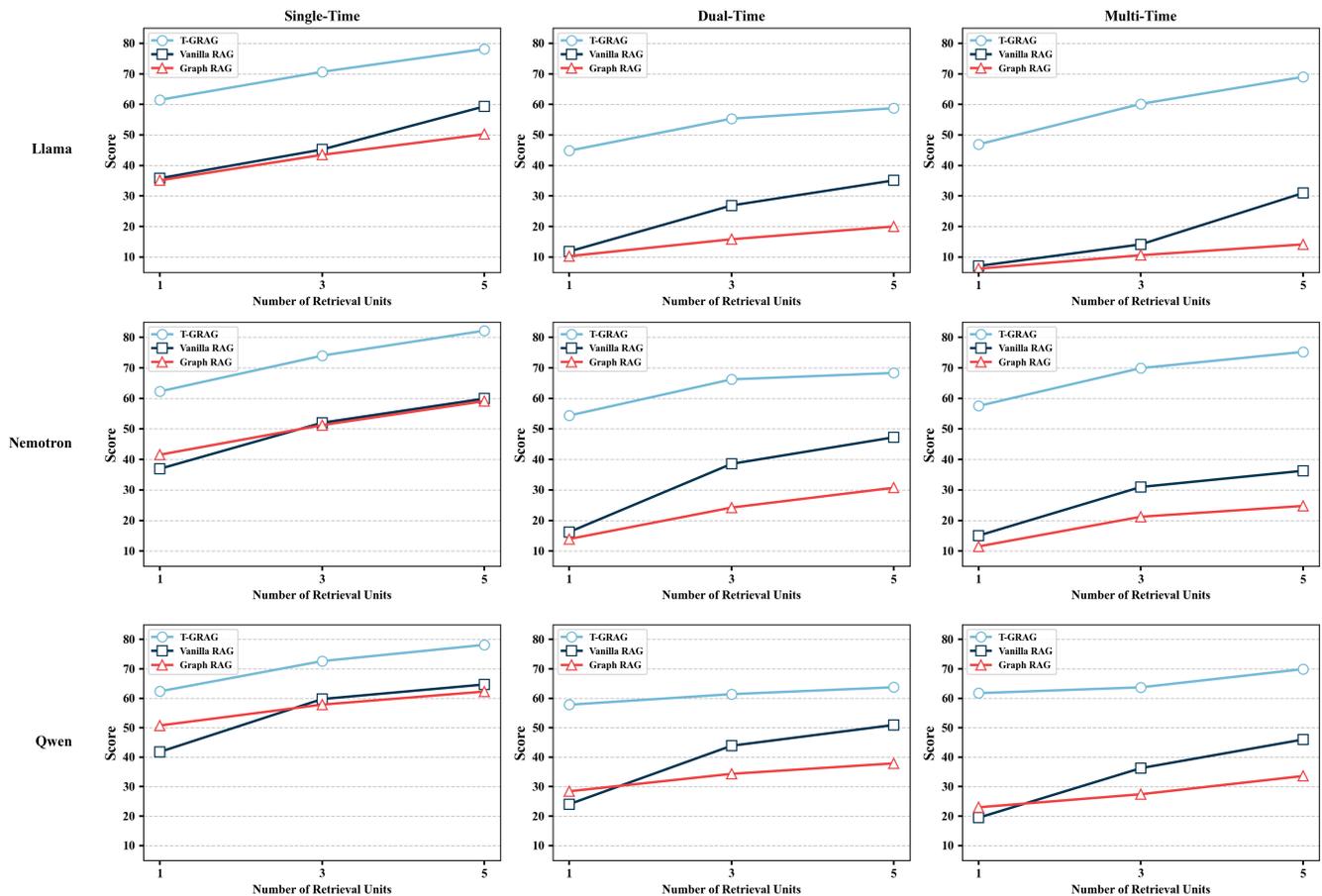

Figure 5: Performance of various RAG methods based on Qwen2.5-72b, Llama-3.1-70B and Llama-3.1-Nemotron-70B under different numbers of retrieval units. Retrieve document parameters of 1, 3, and 5 respectively. We tested Vanilla RAG, Graph RAG and T-GRAG on questions $Q_{Single}$, $Q_{Dual}$ and $Q_{Multi}$.

Table 2: Under identical hardware and model configurations, the average inference time per query for T-GRAG and the baseline GraphRAG.

| Model | Backbone | Single-Time | Dual-Time | Multi-Time | Non-Time |
|---|---|---|---|---|---|
| VanillaRAG | Llama-70b | 9.2 | 10.6 | 10.5 | 9.9 |
|  | Llama-8b | 1.3 | 1.5 | 1.6 | 1.2 |
| GraphRAG | Llama-70b | 14.3 | 15.6 | 15.4 | 14.8 |
|  | Llama-8b | 3.4 | 3.6 | 3.6 | 3.2 |
| T-GRAG | Llama-70b | 12.7 | 13.3 | 14.7 | 13.5 |
|  | Llama-8b | 2.4 | 2.8 | 3.1 | 2.6 |

nodes — can refine the retrieved context. We compare (a) a pipeline using only $\mathcal{R}_{time}$ and $\mathcal{R}_{node}$, and (b) a complete T-GRAG pipeline that also incorporates $\mathcal{R}_{knowledge}$. The results in Figure 4(b) show an average accuracy improvement of 5.81%. This indicates that the multi-level interactive retrieval mechanism helps reduce the interference of redundant node information by refining the granularity of retrieved content. Even in non-temporal scenarios, $\mathcal{R}_{knowledge}$ improves the model's ability to extract relevant evidence from complex knowledge graphs.

**Effect of Temporal Query Decomposition (TQD).** To evaluate the effectiveness of Temporal Query Decomposition (TQD) in addressing multi-temporal constraint questions, we compare two variants of T-GRAG: one with TQD enabled and one without. The only difference between the two settings is the inclusion of the TQD module. As shown in Figure 4(c), enabling TQD leads to a performance gain of 5.79% on dual-time queries ($Q_{Dual}$) and 25.07% on multi-time queries ($Q_{Multi}$). These results demonstrate the significance of TQD in improving model accuracy when handling temporally complex questions. We attribute this improvement to TQD's ability to decompose questions into sub-queries with simplified temporal scopes, which not only reduces temporal reasoning complexity but also avoids retrieval conflicts caused by inconsistent time points. Moreover, TQD helps narrow the size of temporal subgraphs, thereby enhancing the relevance and precision of subsequent node and knowledge retrieval.

**Impact of Retrieved Unit Count.** To investigate the sensitivity of model performance to the number of retrieved units, we vary the retrieval configuration from 1000*5 (default) to 1000*3 and 1000*1. We evaluate three backbone LLMs: Qwen2.5-72B-Instruct, LLaMA3.1-70B-Chat, and Nemotron-70B. The corresponding results are presented in Figure 5. As the number of retrieved units decreases, model performance declines across all methods. Specifically, T-GRAG shows an average accuracy drop of 20.66%, GraphRAG drops by 26.65%, and Vanilla RAG suffers the most with a 54.55% reduction. Notably, T-GRAG and GraphRAG exhibit greater robustness to reduced retrieval context, particularly on temporally demanding questions such as $Q_{Dual}$ and $Q_{Multi}$. This suggests that graph-based retrieval strategies offer better resilience under constrained retrieval conditions. We hypothesize that this advantage stems from the finer granularity of retrieval: while GraphRAG retrieves at the node level, T-GRAG further narrows the scope to knowledge-level granularity within nodes, ensuring that critical information is retained even with fewer retrieved segments.

## 5 Conclusion

In this paper, we propose Temporal GraphRAG (T-GRAG), a novel retrieval-augmented generation framework tailored for dynamic knowledge bases. T-GRAG addresses key challenges such as temporal drift and knowledge redundancy, which often lead to retrieval errors in evolving corpora. Our framework comprises five plug-and-play modules that collaboratively enable temporally-aware indexing, retrieval, and generation. Experiments on our newly constructed Temporal-Dataset, consisting of long-form corporate reports with temporal evolution, show that T-GRAG consistently outperforms GraphRAG and Vanilla RAG on long-text temporal QA tasks. Notably, our Three-layer Interactive Retriever effectively filters redundant knowledge, while the temporal query decomposition module ensures accurate cross-time retrieval. These results demonstrate that T-GRAG significantly improves both retrieval precision and answer accuracy, laying a strong foundation for adaptive, temporally-aware RAG systems. We believe this work underscores the importance of incorporating temporal reasoning into retrieval-based generation and opens new avenues for research in time-sensitive knowledge modeling.

## 6 Acknowledgement

This work is Supported by the National Key R&D Program of China (2023YFC3305102) and Key Laboratory of Cognitive Intelligence and Content Security, Ministry of Education (Grant no.RZZN202411). We extend our gratitude to the anonymous reviewers for their insightful feedback, which has greatly contributed to the improvement of this paper.

## A  Experiments

### A.1  Prompt for LLM Evaluation

---

**Prompt for LLM Evaluation**

**Task Overview:**
   You are tasked with evaluating user answers based on a given question, reference answer, and additional reference text. Your goal is to assess the correctness of the user answer using a specific metric.

**Evaluation Criteria:**
   (1) **Yes/No Questions:** Verify if the user's answer aligns with the reference answer in terms of a "yes" or "no" response.
   (2) **Short Answers/Directives:** Ensure key details such as numbers, specific nouns/verbs, and dates match those in the reference answer.
   (3) **Abstractive/Long Answers:** The user's answer can differ in wording but must convey the same meaning and contain the same key information as the reference answer to be considered correct.

**Evaluation Process:**
   (1) Identify the type of question presented.
   (2) Apply the relevant criteria from the Evaluation Criteria.
   (3) Compare the user's answer against the reference answer accordingly.
   (4) Consult the reference text for clarification when needed.
   (5) Score the answer with a binary label 0 or 1, where 0 denotes wrong and 1 denotes correct.
   (6) **Note:** If the user answer is 0 or an empty string, it should get a 0 score.

**Real Data:**
**Question:** {question}
**User Answer:** {sys_ans}
**Reference Answer:** {ref_ans}
**Reference Text:** {ref_text}
**Output:**
   **Evaluation Form (score ONLY):**
   - Correctness: {score}

---

## B  Prompt for Generating Knowledge Graph

**Entity Extraction Prompts**

**Task:** Given a text document that is potentially relevant to this activity and a list of entity types, identify all entities of those types from the text and all relationships among the identified entities.

**Steps:**

(1) Identify all entities. For each identified entity, extract the following information:
 - **entity_name:** Name of the entity, capitalized
 - **entity_type:** One of the following types: {entity_types}
 - **entity_description:** Comprehensive description of the entity's attributes and activities
 
 Format each entity as "entity"⟨tuple_delimiter⟩ <entity_name> ⟨tuple_delimiter⟩ <entity_type> ⟨tuple_delimiter⟩ <entity_description>.

(2) From the entities identified in step 1, identify all pairs of (source_entity, target_entity) that are *clearly related* to each other. For each pair of related entities, extract the following information:
 - **source_entity:** name of the source entity, as identified in step 1
 - **target_entity:** name of the target entity, as identified in step 1
 - **relationship_description:** explanation as to why you think the source entity and the target entity are related
 - **relationship_strength:** A numeric score showing the strength of the relationship between the source and target entities
 
 Format each relationship as "relationship"⟨tuple_delimiter⟩<source_entity>⟨tuple_delimiter⟩<target_entity>⟨tuple_delimiter⟩<relationship_description>⟨tuple_delimiter⟩<relationship_strength>.

(3) Return output in English as a single list of all the entities and relationships identified in steps 1 and 2. Use **{record_delimiter}** as the list delimiter.

(4) When finished, output **{completion_delimiter}**.

**Examples:**

Entity_types: [person, technology, mission, organization, location]

Text: Alex's frustration was overshadowed by Taylor's certainty. Their competitive tension kept him alert, as his and Jordan's commitment to discovery felt like a silent rebellion against Cruz's controlling vision. Unexpectedly, Taylor paused beside Jordan, observing the device with reverence. "If we can understand this tech, it could change the game for all of us," Taylor said quietly. Their earlier dismissal had shifted to reluctant respect, and a brief, silent truce formed between Jordan and Taylor. This small but noticeable change was acknowledged by Alex, who knew they had all arrived here through different paths.

**Output:**
- ("entity"⟨tuple_delimiter⟩"Alex"⟨tuple_delimiter⟩"person"⟨tuple_delimiter⟩"Alex is a character who experiences frustration and is observant of the dynamics among other characters.")
- ("entity"⟨tuple_delimiter⟩"Taylor"⟨tuple_delimiter⟩"person"⟨tuple_delimiter⟩"Taylor is portrayed with authoritarian certainty and shows a moment of reverence towards a device, indicating a change in perspective.")
- ("entity"⟨tuple_delimiter⟩"Jordan"⟨tuple_delimiter⟩"person"⟨tuple_delimiter⟩"Jordan shares a commitment to discovery and has a significant interaction with Taylor regarding a device.")
- ("entity"⟨tuple_delimiter⟩"Cruz"⟨tuple_delimiter⟩"person"⟨tuple_delimiter⟩"Cruz is associated with a vision of control and order, influencing the dynamics among other characters.")
- ("entity"⟨tuple_delimiter⟩"The Device"⟨tuple_delimiter⟩"technology"⟨tuple_delimiter⟩"The Device is central to the story, with potential game-changing implications, and is revered by Taylor.")
- ("relationship"⟨tuple_delimiter⟩"Alex"⟨tuple_delimiter⟩"Taylor"⟨tuple_delimiter⟩"Alex is affected by Taylor's authoritarian certainty and observes changes in Taylor's attitude towards the device."⟨tuple_delimiter⟩7)
- ("relationship"⟨tuple_delimiter⟩"Alex"⟨tuple_delimiter⟩"Jordan"⟨tuple_delimiter⟩"Alex and Jordan share a commitment to discovery, which contrasts with Cruz's vision."⟨tuple_delimiter⟩6)
- ("relationship"⟨tuple_delimiter⟩"Taylor"⟨tuple_delimiter⟩"Jordan"⟨tuple_delimiter⟩"Taylor and Jordan interact directly regarding the device, leading to a moment of mutual respect and an uneasy truce."⟨tuple_delimiter⟩8)
- ("relationship"⟨tuple_delimiter⟩"Jordan"⟨tuple_delimiter⟩"Cruz"⟨tuple_delimiter⟩"Jordan's commitment to discovery is in rebellion against Cruz's vision of control and order."⟨tuple_delimiter⟩5)
- ("relationship"⟨tuple_delimiter⟩"Taylor"⟨tuple_delimiter⟩"The Device"⟨tuple_delimiter⟩"Taylor shows reverence towards the device, indicating its importance and potential impact."⟨tuple_delimiter⟩9)

**Real Data:**

Entity types: {entity_types}

Text: {input_text}

Output:

## B.1 Prompt for Temporal Query Decomposition

**TQD Prompts**

**Task:** Break down a question containing multiple years into individual questions for each year. Each question should include the year and the specific question related to that year.
   !!! output format must like [<year><SEP><question>]!!!
   !!! output format must like [<year><SEP><question>]!!!
   !!! output format must like [<year><SEP><question>]!!!
   !!!Do not output "Here is the summary of the provided text" Or similarly, directly output the output!!!
   !!!Do not output "Here is the summary of the provided text" Or similarly, directly output the output!!!
   !!!Do not output many answer, just choose one, directly output the output!!!
   the output only is [<year><SEP><question>], Do not have any other text.
**Examples:**
   **Example 1:** *question:* "How many people worked for the Audi Group worldwide in 2017 and 2023, respectively?" *output:* [2017<SEP>"How many people worked for the Audi Group worldwide in 2017"] [2023<SEP>"How many people worked for the Audi Group worldwide in 2023"]
   **Example 2:** *question:* "How much CO2 was saved through Audi's Aluminum Closed Loop process in 2019 compared to the Aluminum Closed Loop Pilot Project in 2017?" *output:* [2019<SEP>"How much CO2 was saved through Audi's Aluminum Closed Loop process in 2019"] [2017<SEP>"How much CO2 was saved through Audi's Aluminum Closed Loop Pilot Project in 2017"]
   **Example 3:** *question:* "What were the energy intensity of Audi's car production in 2020, 2019 and 2017 respectively?" *output:* [2020<SEP>"What was the energy intensity of Audi's car production in 2020?"] [2019<SEP>"What was the energy intensity of Audi's car production in 2019?"] [2017<SEP>"What was the energy intensity of Audi's car production in 2017?"]
   **Example 4:** *question:* "How does Audi's electric vehicle model launch plan in 2023-2025 compare to its electrification efforts in 2021-2022?" *output:* [2023-2025<SEP>"How does Audi's electric vehicle model launch plan in 2023-2025?"] [2021-2022<SEP>"How does Audi's electrification efforts in 2021-2022?"]
   **Real Data:**
   question:{question}
   output:

## B.2 Prompt for LLM Enhancement Generation

---
**Subquestion Enhancement Generation Prompts**

**Role:**
You are a helpful assistant who can answer questions about the data in the provided table.
**Goal:**
Use the relevant data provided in the table to answer questions about the data in the table. If you don't know the answer, just say 'I'm sorry I don't know the answer' directly.
**Output format:**
The output needs to be concise, which is the conclusion of your final answer to this question. Do not output the thought process.
**Question:** {question}
**Data tables:** {context_data}

---

**Final Question Answer Prompts**

**Role:**
   You now need to answer this question based on the completed sub questions and corresponding answers I have provided to you. I have broken down the question you need to answer into multiple different sub questions and generated corresponding answers. You need to answer this question based on the completed sub questions and corresponding answers. If you don't know the answer, just say so. Do not make anything up.
   **Output format:**
   The output needs to be concise, which is the conclusion of your final answer to this question. Do not output the thought process
   **Input Data:**
   **final question:** {question}
   **sub questions and corresponding answers:** {qa_dict}

---

## C  Prompts for Building Dataset

The prompts designed to build dataset can be categorized into three distinct types: prompt for generating summary, prompt for generating key points, and prompt for creating QA pairs.

## C.1  Prompt for Generating Summary

---
**Prompt for Generating Summary from chunks**

You are an AI assistant tasked with reading and understanding a lengthy text and generating a summary in pure text form. The summary should adhere to the following specific requirements:
   1. The summary should cover a broad range of information, aiming to include the vast majority of details from the original text.
   2. Minimize the use of pronouns and clearly specify the names of entities to ensure clarity.
   3. Retain any time-related data and other crucial details without omitting them, ensuring the summary is comprehensive and accurate.
   4. Avoid using quotation marks or other special symbols, and present the summary in plain text without complex formatting or lists, structuring it into standard paragraphs.
   Please ensure the output aligns with these requirements.

## C.2  Prompt for Generating Key Points

**Prompt for Generating Key Points from Summary**

**Task:**
You are an artificial intelligence assistant, please extract key points from the article according to the following rules:
1. The key points should be independent of each other and the content should avoid overlapping as much as possible.
2. Key points should be concise, accurate, and complete, especially when it comes to numbers, names, and dates.
3. The key points should not have complex formats or line breaks, just one or two sentences
4. If the key points do not involve events that occurred in year, please ignore them and keep only discussing events that occurred in year.
5. Basically, pronouns such as "he, she, them, it" cannot be used, and it is necessary to clearly indicate the entity you are referencing in the key points.
6. The following opening phrases are not allowed: -The article discussed -The article shows -The article emphasizes -The speaker said -The author mentioned... and so on.

**Output Format:**
The response should be JSON formatted as follows:
{"point-id":"point"}
The beginning and end of the answer must be {}
!!!The answer is only in JSON format,Don't output here is the output Or similarly, directly output the output!!!

**Example:**
**Input:** Here is the summary of the provided text in plain text, adhering to the specified requirements:The Audi Corporate Responsibility Report 2012, published by AUDI AG, presents the company's work in corporate responsibility (CR) to stakeholders and the public for the first time. As a member of the UN Global Compact since February 2012, AUDI AG adheres to the ten principles in Human Rights, Labor, Environment, and Anti-Corruption. The report covers the period from January 1 to December 31, 2012, and includes supplementary information up to the editorial deadline in March 2013.

The Audi Group, comprising the Audi and Lamborghini brands, is a leading carmaker in the premium and supercar segment. In 2012, the group expanded its portfolio by acquiring DUCATI MOTOR HOLDING S.P.A., entering the motorcycle market. The company also manufactures engines in Győr, Hungary, for Audi, other Volkswagen Group companies, and third parties.The report is structured around five core themes: Operations, Product, Environment, Employees, and Society. It conforms to the G3.1 Guidelines of the Global Reporting Initiative (GRI) and the Automotive Sector Supplement, with an Application Level confirmed as B+. An independent audit was conducted by PricewaterhouseCoopers.Audi's Board of Management, including Prof. Rupert Stadler, Luca de Meo, Dr. Frank Dreves, Wolfgang Dürheimer, Dr. Bernd Martens, Prof. h. c. Thomas Sigi, and Axel Strotbek, emphasizes the importance of sustainability and corporate responsibility. The company aims to ensure a livable future for generations to come, inspired by its philosophy of V̇orsprung durch Technik.The report highlights Audi's product range, including various models produced in different locations: Neckarsulm and Ingolstadt, Germany (e.g., A4 Sedan, A6 Sedan, A8, R8 Coupé); Martorell, Spain (Q3 by SEAT, S.A.); Sant'Agata Bolognese, Italy (Lamborghini models like Gallardo Coupé and Aventador Coupé); and Bologna, Italy (Ducati motorcycles such as Diavel, Hypermotard, and Superbike).Key figures and data for the period 2010-2012 are included in the report, which can be viewed in its entirety online in German and English at www.audi.com/cr-report2012. The next fully revised report will be published in the first half of 2015, with main key figures for 2013 revised in the first half of 2014. Readers can contact Dr. Peter F. Tropschuh, Head of Corporate Responsibility at AUDI AG, with questions or comments.

**Output:**
"point-1": "AUDI AG published its first Corporate Responsibility Report in 2012, covering the period from January 1 to December 31, 2012, with supplementary information up to March 2013."
"point-2": "AUDI AG became a member of the UN Global Compact in February 2012, adhering to ten principles in Human Rights, Labor, Environment, and Anti-Corruption."
"point-3": "The Audi Group, including Audi and Lamborghini brands, acquired DUCATI MOTOR HOLDING S.P.A. in 2012 and entered the motorcycle market."
"point-4": "The report is structured around five core themes: Operations, Product, Environment, Employees, and Society, and conforms to the G3.1 Guidelines of the Global Reporting Initiative (GRI) with an Application Level of B+."
"point-5": "The report includes an independent audit conducted by PricewaterhouseCoopers."
"point-6": "Audi's Board of Management in 2012 emphasized sustainability and corporate responsibility, aiming to ensure a livable future for future generations."
"point-7": "In 2012, Audi produced models in Germany (Neckarsulm, Ingolstadt), Spain (Martorell), Italy (Sant'Agata Bolognese, Bologna), including various car and motorcycle models."
"point-8": "Key figures and data for 2010-2012 are available in the report, which can be accessed online at www.audi.com/cr-report2012. The next fully revised report is scheduled for release in 2015."

**Real Data:**
**Input:** {summary}
**Output:**

## C.3 Prompt for Creating QA pairs

---
**Prompt for Creating Single-Time QA pairs (remove time constraint as for Non-Time QA)**

Please generate ten questions and their answers based on the input text.
**Specific Requirements:**
   1. The questions should be diverse, covering different angles such as people, numbers, places, etc.
   2. The information in the questions should be rich and clear enough to avoid ambiguity in the answers.
   3. Design three challenging questions, avoiding simple string matching.
   4. Each question's answer should be concise, avoiding redundancy or repetition of the information in the question.
   5. The events included in the questions must have clear time attributes as specified in the original text. If the event's time cannot be determined, abandon that question and find another event to create a question.(very important)
   6. Each question must contain a time attribute. In other words, the question must include a specific time reference.(very important)
**Output Format:**
   The output format should be in JSON format:
   { "Question": "Question",
   "Answer": "Answer",
   "OriginalText": "Original content of the text (keep the original content and format)"}
**OriginalText:** {text}

---
**Prompt for Creating Dual-Time QA pairs (increase time points as for Multi-Time QA )**

You are an artificial intelligence assistant, and I now need you to help me generate one time-series QA pairs. I will provide you with the key points and corresponding original texts of two similar events from two annual reports.
**Specific Requirements:**
   1. The question must be answered using the original texts corresponding to these two timestamps, and there should be temporal comparability between these original texts.
   2. The questions should be close-ended.
   3. The question needs to be an inquiry about specific entities, numbers, or time, not about abstract concepts (such as Audi's strategy, Audi's plans in the field of electric vehicles)
   4. The information in the question should be sufficient and clear, avoiding ambiguous answers.
   5. Multiple sub questions cannot be included in the problem, and the problem should be clear and specific.
   6. The problem must contain two timestamps, year1 and year2.
   7. The answer should be concise and clear, avoiding being lengthy or repetitive.
**Output Format:**
   The output format should be in JSON format:
   { "Question": "Question",
   "Answer": "Answer",
   "Original text from {year1} report": "<original text from {year1} report>",
   "Original text from {year2} report": "<original text from {year2} report>"}
**Real Data:**
   **Input Data:**
   keypoint in {year1} Annual Report:{keypoint1}, Corresponding original text: {text1};
   keypoint in {year2} Annual Report:{keypoint2}, Corresponding original text: {text2}
   **Output:**

---